\documentclass[sigconf]{acmart}
\renewcommand\footnotetextcopyrightpermission[1]{} 
\AtBeginDocument{%
  \providecommand\BibTeX{{%
    \normalfont B\kern-0.5em{\scshape i\kern-0.25em b}\kern-0.8em\TeX}}}

\setcopyright{acmcopyright}
\copyrightyear{2021}
\acmYear{2021}
\acmDOI{10.1145/1122445.1122456}

\acmSubmissionID{212}

\usepackage{hyperref}       
\usepackage{url}            
\usepackage{booktabs}       
\usepackage{amsfonts}       
\usepackage{nicefrac}       
\usepackage{microtype}      

\usepackage{url}
\usepackage{boxedminipage}

\usepackage{amsmath,amsfonts,bm}

\def\eqref#1{equation~\ref{#1}}

\def\1{\bm{1}}

\DeclareMathAlphabet{\mathsfit}{\encodingdefault}{\sfdefault}{m}{sl}
\SetMathAlphabet{\mathsfit}{bold}{\encodingdefault}{\sfdefault}{bx}{n}

\usepackage{hyperref}
\usepackage{wrapfig,lipsum,booktabs}
\usepackage{amsmath}
\usepackage{cleveref}
\usepackage{booktabs}
\usepackage{bm}
\usepackage{algorithm}
\usepackage{algpseudocode}
\usepackage{makecell}
\usepackage{graphicx}
\usepackage{nccmath}
\usepackage{caption}
\usepackage{multirow}
\usepackage{makecell}
\usepackage{subcaption}
\usepackage{tabularx}
\usepackage{colortbl}
\newcommand{\B}[1] {\boldsymbol{#1}}

\def\bm{{\B{m}}}

\title{Make Lead Bias in Your Favor: Zero-shot Abstractive News Summarization}

\author{Chenguang Zhu}
\email{chezhu@microsoft.com}
\affiliation{%
  \institution{Microsoft Cognitive Services Research Group}
}

\author{Ziyi Yang}
\email{zy99@stanford.edu}
\affiliation{%
 \institution{Stanford University}
}

\author{Robert Gmyr, Michael Zeng, Xuedong Huang}
\email{{rogmyr,nzeng,xdh}@microsoft.com}
\affiliation{%
  \institution{Microsoft Cognitive Services Research Group}
}

\makeatletter
\newcommand{\thickhline}{%
    \noalign {\ifnum 0=`}\fi \hrule height 1pt
    \futurelet \reserved@a \@xhline
}
\makeatother

\begin{document}
\begin{abstract}
Lead bias is a common phenomenon in news summarization, where early parts of an article often contain the most salient information. While many algorithms exploit this fact in summary generation, it has a detrimental effect on teaching the model to discriminate and extract important information in general. We propose that the lead bias can be leveraged in our favor in a simple and effective way to pre-train abstractive news summarization models on large-scale unlabeled news corpora: predicting the leading sentences using the rest of an article.
We collect a massive news corpus and conduct data cleaning and filtering via statistical analysis. We then apply the proposed self-supervised pre-training to existing generation models BART and T5 for domain adaptation.
Via extensive experiments on six benchmark datasets, we show that this approach can dramatically improve the summarization quality and achieve state-of-the-art results for zero-shot news summarization without any fine-tuning. For example, in the DUC2003 dataset, the ROUGE-1 score of BART increases 13.7\% after the lead-bias pre-training. We deploy the model in Microsoft News and provide public APIs as well as a demo website for multi-lingual news summarization.
\end{abstract}



\maketitle
\pagestyle{plain}

\section{Introduction}
\begin{figure*}[t]
\centering
  \includegraphics[width=15cm]{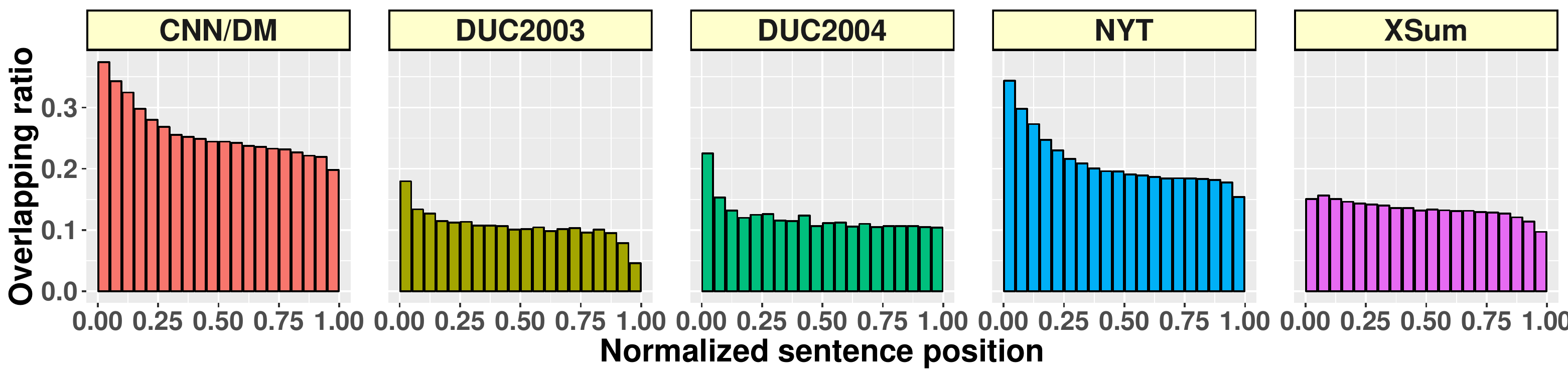}
  \caption{Average overlapping ratio of words between an article sentence and the reference summary, grouped by the normalized position of the sentence (the size of bin is 0.05). The ratio is computed on the training set of corresponding datasets.}
  \label{fig:showleadbias}
\end{figure*}

The goal of text summarization is to condense a piece of text into a shorter version that contains the salient information. Due to the prevalence of news articles and the need to provide succinct summaries for readers, a majority of existing datasets for summarization come from the news domain \cite{cnn,nyt,tconvs2s}. However, according to journalistic conventions, the most important information in a news report usually appears near the beginning of the article \cite{top3,top3_2}, known as the inverted pyramid structure. For example, Figure~\ref{fig:showleadbias} displays the average ratio of words in an article sentence that also appear in the human-written reference summary, grouped by the normalized sentence position in the article. As shown, in most news corpora, sentences appearing at the beginning of the article have the highest average overlapping ratio with the reference summary, corroborating the lead bias in news articles.
 
While it facilitates faster and easier understanding of the news for readers, this lead bias can cause undesirable consequences for summarization models. The output from these models is inevitably affected by the positional information of sentences. As a result, the models find it hard to discriminate and extract important information for text that does not demonstrate this bias. 
For instance, \cite{countertop} discovers that most models' performances drop significantly when a random sentence is inserted in the leading position, or when the sentences in a news article are shuffled.

Additionally, most current summarization models are fully supervised and require time-consuming and labor-intensive annotations to feed their insatiable appetite for labeled data. For example, the CNN/DailyMail dataset \cite{nyt} contains 313k articles, where the summaries are written by editors. Therefore, some recent work \cite{domaintransfer} leverage domain transfer to apply summarization models trained on one dataset to another dataset. But this method may be affected by the domain drift problem and still suffers from the lack of labelled data.

The recent promising trend of pre-trained language models \cite{bert, gpt} proves that massive data can be used in a self-supervised fashion to boost the performance of NLP models. But it remains a challenge how to design pre-training goals for text summarization. In this paper, we put forward a novel method to leverage the lead bias of news articles in our favor to conduct large-scale pre-training of summarization models. The idea is to predict the leading sentences of a news article given the rest of the content. This immediately renders the large quantity of unlabeled news corpora available for building summarization models.

Thus, we collect a massive corpus of online news articles over three years from the index of the Bing search engine. We then conduct thorough data cleaning and filtering to ensure the quality of leading sentences as the delegate summary. We compute the overlapping ratio of non-stopping words between the top 3 sentences and the rest of the article. As a higher ratio implies a closer semantic connection, we only keep articles for which this ratio is higher than a threshold determined via statistical analysis. We find that in the CNN/Daily Mail dataset, the median of this overlapping ratio between the reference summary and the article is as much as 0.841. And the median of the ratio between the leading 3 sentences and the rest of article is 0.471. Thus, we choose a threshold ratio of 0.65 for our data to strike a balance between the quality of leading sentences and the size of training data. The retained data for pre-training contains 21.4M news articles and has a median overlapping ratio of 0.734.

Inspired by the effectiveness of additional pre-training to adapt language models to downstream domains, we start from the pre-trained language generation models BART \cite{bart} and T5 \cite{t5} and continue to pre-train them on our lead prediction task.
Thus, our pre-training can be considered as domain adaptation.
The resulting models BART-LB and T5-LB are leveraged in a zero-shot fashion, i.e. directly applied to target tasks without accessing \textit{any} information for fine-tuning.
Therefore, the same pre-trained model can be used across various news summarization tasks.

We conduct extensive evaluations on six news summarization datasets. Results show that our models significantly improve the summary quality over the original BART and T5, outperforming other zero-shot and unsupervised summarization baselines. For example, BART-LB outperforms BART$_\text{LARGE}$ by 13.7\%, 8.3\% and 6.9\% in ROUGE-1 on DUC2003, DUC2004 and XSum. Also, BART-LB outperforms the zero-shot version of the state-of-the-art summarization model PEGASUS \cite{pegasus} by 7.6\%, 6.9\%, and 1.8\% in ROUGE-1 on CNN/DailyMail, XSum and Gigawords, respectively.

We then provide insights from various aspects to analyze the results:

\begin{enumerate}
    \item Via ratio of novel $n$-grams, we show that our model does not learn to simply copy leading sentences from the article.
    \item Our pre-training can mostly improve the quality of summaries of medium and long lengths.
    \item Our models BART-LB and T5-LB are not sensitive to decoding hyper-parameters like maximum summary length and beam width.
\end{enumerate}
    
We conduct a human evaluation and show that the lead-bias pre-training can improve both readability and relevance of generated summaries. We also showcase examples of summaries and conduct error analysis to investigate the common types of error in the summaries generated by our model.

Finally, we deploy the pre-trained model in Microsoft News coupled with translation and text-to-speech technologies. We provide public APIs as well as a demo website to support news summarization in 30 different languages with both text and speech output.


\section{Related work}
\label{Sec:rw}
\subsection{Unsupervised Text Summarization} 
Traditional abstractive summarization models \cite{pgnet,drm} require article-summary pairs as the training data. The model is optimized via the discrepancy between the generated summary and the ground-truth summary, e.g. cross entropy. However, the reference summaries require lots of manual work from experts. Therefore, unsupervised abstractive summarization (UAS) models aim to learn to summarize from articles alone. Among these approaches, SEQ$^3$ \cite{baziotis2019seq} tries to reconstruct the input article from the generated summary. Summaries that can be used to better generate back the original article are preferred. MeanSum \cite{meansum} applies the same reconstruction idea to multi-document summarization. \cite{fevry} adopts denoising autoencoders for sentence compression. Brief \cite{wang2018learning} leverages reinforcement learning and adversarial training to improve the readability of generated summaries.
SummAE \cite{summae} projects articles and sentences into a common space from which reconstruction can be conducted. TED \cite{ted} employs theme modeling and a denoising autoencoder to enhance the quality of summaries. Although TED also leverages lead bias for pre-training, it is trained from scratch and is employed in unsupervised summarization, while our work employs lead bias for domain adaptation and is used for zero-shot summarization.

In general, unsupervised summarization models need to be trained on the articles in the training set of target tasks, and the model trained on one dataset may not be suitable on another dataset due to domain shift. In comparison, our model is strictly zero-shot in that it does not even access articles in the training set of downstream tasks. And the same model instance can be applied to multiple news summarization tasks.

\subsection{Pre-trained generation models} 
In recent years, pre-trained language models have proved to be quite helpful in natural language generation tasks. UniLM \cite{unilm} consolidates language understanding and generation into a single Transformer architecture. BART \cite{bart} is a generation model pre-trained by denoising to recover corrupted text. T5 \cite{t5} presents a systematic study over pre-training objectives, architectures, datasets, and other factors for pre-trained language generation models. Built upon large-scale corpora, these models employs self-supervised learning such as denoising autoencoder and masked language model to learn effective semantic representations for language generation.

In theory, any generation model can be directly used for zero-shot abstractive summarization (ZAS). 
For example, the GPT-2 model \cite{gpt} can produce summaries given an article appended with \textit{TL;DR:}. PEGASUS \cite{pegasus} uses gap sentences generation (GSG) and masked language modeling (MLM) as pre-training objectives to build a summarization model. Although GSG includes top-$m$ sentences as decoder targets, we apply a novel news-lead cleaning and filtering mechanism to the pre-training data and the resulting models significantly outperform PEGASUS in multiple datasets. \cite{zou2020pretraining} uses sentence reordering, next sentence generation and masked document generation for summarization pre-training. But as the pre-training is not directly related to summarization, the model requires further fine-tuning to be applied in downstream tasks.


\section{Pre-training with Lead Bias}
\label{Sec:pre}
\subsection{Lead bias}
News articles usually follow the convention of placing the most important information early in the content, forming an inverted pyramid structure \cite{top3,top3_2,countertop}. As a result, the lead baseline, which simply takes the top few sentences as the summary, can achieve a rather strong performance in news summarization. For example, the Lead-3 baseline, which takes the first 3 article sentences as the summary, can get a higher ROUGE score than many deep learning based models in CNN/DailyMail and NYT datasets \cite{cnn,nyt}. This positional bias brings lots of difficulty for models to extract salient information from the article. For instance, \cite{countertop} discovers that most models' performances drop significantly when a random sentence is inserted in the leading position, or when the sentences in a news article are shuffled.

On the other hand, news summarization, just like many other supervised learning tasks, suffers from the lack of labeled training data. Abstractive summarization is especially data-hungry since the efficacy of models depends on high-quality handcrafted summaries.

We propose that the lead bias in news articles can be leveraged in our favor to train an abstractive summarization model without labeled summaries. Given a news article, we treat the top three sentences, denoted by Lead-3, as the target summary, and use the rest of the article as news content. The goal of the summarization model is to produce Lead-3 using the rest of the content.

We note that this idea of utilizing structural bias for large-scale summarization pre-training is not limited to specific types of models. It can be applied to other types of text as well: academic papers with abstracts, novels with editor's notes, books with tables of contents.

\subsection{Data collection and filtering}
We collect three years of public online news articles from June 2016 to June 2019 via the index of the Bing Search Engine. We filter out articles that overlap with any of the evaluation datasets\footnote{Among the six benchmark news summarization datasets used in experiments, five were published before 2016. Only XSum has 1 year overlap with our data range, and we remove the overlapping articles from our pre-training dataset.}.

However, one should carefully examine and clean the source data to take advantage of lead bias, as the top three sentences may not always form a good summary.

First, to ensure that the summary is concise and the article contains enough salient information, we only keep articles with 10-150 words in the top three sentences and 150-1200 words in the rest, and that contain at least 6 sentences in total. In this way, we filter out i) articles with excessively long content to reduce memory consumption; ii) very short leading sentences with little information, which are unlikely to be a good summary. To encourage the model to generate abstractive summaries, we also remove articles where any of the top three sentences is exactly repeated in the rest of the article.

\begin{figure}[t]
    \centering
\includegraphics[width=0.45\textwidth]{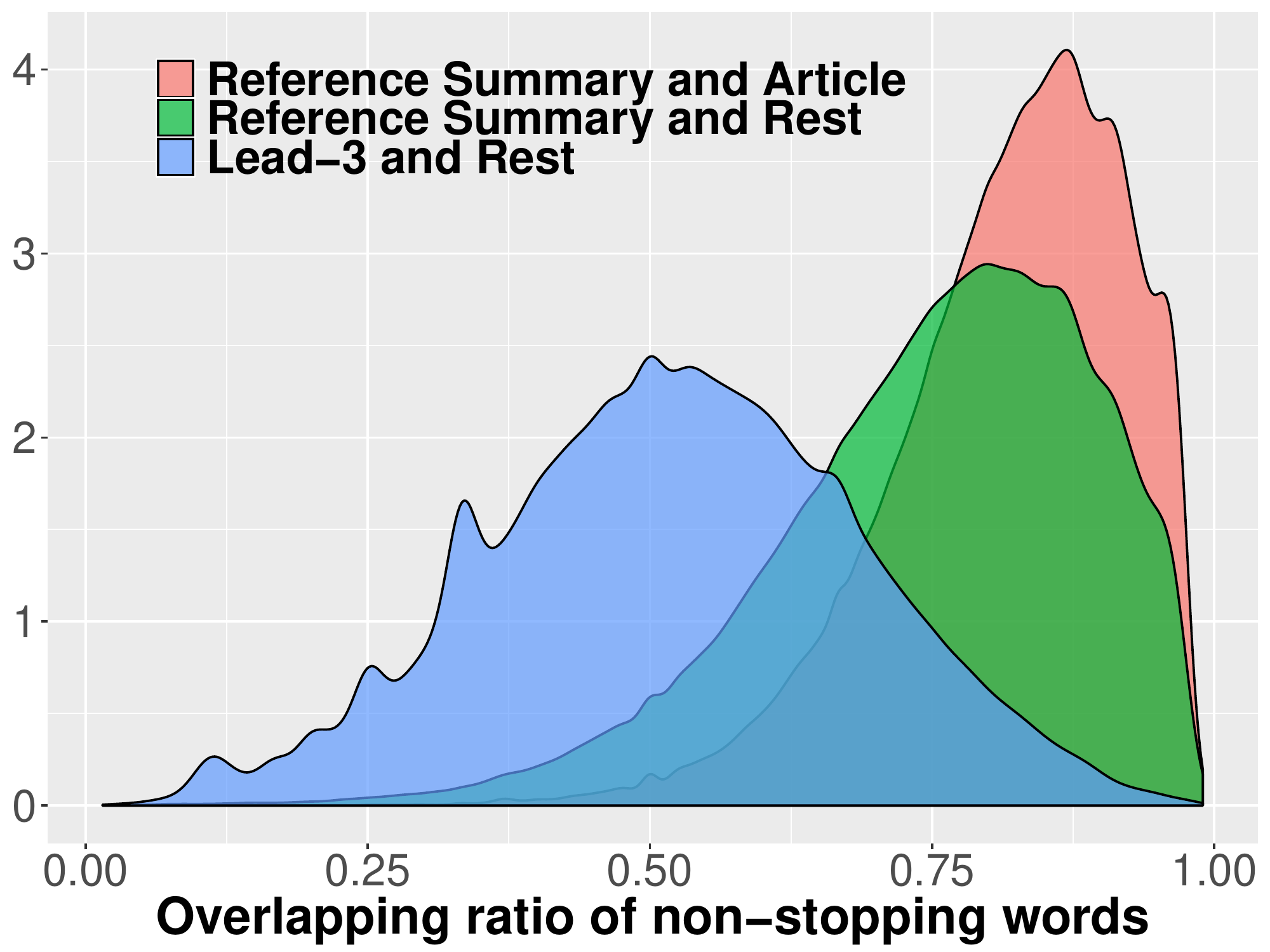}
\caption{The distribution of the overlapping ratio of non-stopping words between: (Red) the reference summary and the article; (Green) the reference summary and the article excluding the first 3 sentences, i.e. Rest; and (Blue) the leading 3 sentences, i.e. Lead-3, and Rest. The area under each curve is 1. All ratios are computed on CNN/DailyMail.}
\label{fig:overlap_ratio}
\end{figure}

Second, we remove articles whose top three sentences may not form a relevant summary. For this purpose, we compute \textit{the portion of non-stopping words} in the top three sentences, i.e. Lead-3, that are also in the rest of an article, denoted by Rest.
A higher portion implies that Lead-3 is more relevant to Rest and has a higher chance of being inferred from Rest. 

To verify this hypothesis, we plot the distribution of the overlapping ratio of non-stopping words between the reference summary and the article in the training set of CNN/DailyMail, as indicated by the red curve in Figure~\ref{fig:overlap_ratio}. This red curve has a median value as high as 0.841. This median drops moderately to 0.778 between the reference summary and Rest (green curve). And the ratio we use for filtering, i.e. that between Lead-3 and Rest, is shown in blue and has a median value of 0.471 in CNN/DailyMail. This indicates that for many articles, Rest has a much lower overlapping ratio with Lead-3 than with the reference summary. So we need to filter out these articles to ensure the quality of Lead-3 as a surrogate summary. 

We end up choosing a threshold of 0.65 for the overlapping ratio between Lead-3 and Rest to strike a balance between the quality of Lead-3 and the size of training data. This filters out 83.2\% of the collected data. The retained data for pre-training has a median overlapping ratio of 0.734 between Lead-3 and Rest.

\begin{table*}[t]
    \centerline{
    \begin{tabular}{l|l|c|c}
    \thickhline
        Datasets & docs (train/val/test) & 
        avg. words in article & avg. words in summary \\ \thickhline
        DUC2003 & 0 / 0 / 624 & 555.6 & 38.2 \\
        DUC2004 & 0 / 0 / 500 & 587.1 & 40.3\\
        XSum & 204,045/11,332/11,334 & 431.1 & 23.3 \\
        CNN/DailyMail & 287,227/13,368/11,490& 683.6 & 48.7 \\
        NYT  & 96,834/4,000/3,452 & 800.0& 45.5 \\
        Gigaword & 3.8M/189k/1951 &  29.0 & 8.8 \\
        \thickhline
    \end{tabular}}
    
    \caption{\label{tab:statistics} Comparison of summarization datasets in the experiments: size of training, validation, and test sets and average document and summary length.
    }  
\end{table*}

\begin{table}[h]
    \centering
    \begin{tabular}{l|c|c|c}
    \thickhline
        \textbf{Dataset} & \textbf{sum. minlen} & \textbf{sum. maxlen} & \textbf{beam width} \\
        \hline
        CNN/DM & 56 & 142 & 4 \\
        NYT  & 56 & 142 & 4 \\
        XSum & 11 & 62 & 6 \\
        DUC2003 & 6 & 26 & 1 \\
        DUC2004 & 6 & 26 & 1\\
        Gigaword & 4 & 24 & 4\\
        \thickhline
    \end{tabular}
    \caption{\label{tab:hyperparam} Hyper-parameters used on each dataset, including minimum/maximum length of produced summary and beam width. These parameters are determined by the implementation of summarization models from previous literature.}
\end{table}

In total, the above methods filter about 95\% of the original set of articles and we end up with 21.4M news articles for pre-training. We also clean the article content by using regular expressions to remove article prefixes containing reporter names, media agencies, dates or other information irrelevant to the content, e.g. ``New York (CNN) --'', ``Jones Smith, May 10th, 2018:''.

In the data for pre-training, the average number of words in Lead-3 is 60.0 and the average number of words in Rest is 602.5. Therefore, the data for pre-training contains 14.2 billion words in total.

\subsection{Pre-training}
Domain adaptation has been shown to be effective to further improve the quality of pre-trained models \cite{gururangan2020don}. Specifically, a language model pre-trained for general purposes should be further pre-trained on data in the same domain as the target tasks to achieve better performance. As the lead bias is specific to news articles, we frame the proposed lead-bias pre-training as domain adaptation.

Thus, we initialize two versions of our model with pre-trained language model BART-Large \cite{bart} and T5-Large \cite{t5} respectively. These two models have been pre-trained for general purpose of language generation. Both models use the seq2seq architecture based on Transformer \cite{transformer}. We then conduct the second phase of pre-training using the lead bias. More implementation details are described in Section~\ref{sec:impl}.

After pre-training on our lead-bias data, we denote the models as \textbf{BART-LB} and \textbf{T5-LB} respectively. As we use the zero-shot setting, the same model instance is evaluated on all summarization datasets.

\section{Experiments}
\label{Sec:exp}
\subsection{Datasets}
We evaluate our model on 6 benchmark news summarization datasets.

\textbf{DUC2003 / DUC2004} \cite{duc} contain 624 and 500 news articles respectively. The articles are from the New York Times and Associated Press Wire services, each paired with 4 human-generated reference summaries. Both DUC2003 and DUC2004 only contain test data.

\textbf{XSum} \cite{tconvs2s} contains 227K BBC news articles from 2010 to 2017. Each article comes with a professionally written single-sentence summary.

\textbf{CNN/DailyMail} \cite{cnn} contains 313k articles from CNN and Daily Mail, where the summary is written by experts shown as bullet points. We use the non-anonymized version as in \cite{pgnet}.

\textbf{The New York Times Annotated Corpus (NYT)} \cite{nyt} contains 110k articles written and published by the New York Times between 1987 and 2007. The summaries are written by library scientists. We follow the filtering procedure of \cite{bertsum} to remove documents with summaries of fewer than 50 words.

\textbf{Gigaword} contains 4M pairs of articles and summaries from news articles from seven publishers \cite{gigaword}. Different from other datasets, the summary is the headline and the article is the first sentence of the original news.


More details of these datasets including test set size, average number of words in the article and summary are shown in Table~\ref{tab:statistics}.

\subsection{Metrics}
We employ the standard ROUGE-1, ROUGE-2 and ROUGE-L metrics \cite{rouge} to evaluate all summarization models. These three metrics respectively evaluate the accuracy on unigrams, bigrams and longest common subsequence. ROUGE metrics have been shown to highly correlate with the human judgment \cite{rouge}. To align with previous results, we report ROUGE recall scores on NYT and report ROUGE F1 scores on the other datasets. In NYT, the generated summary is truncated to the same length as the ground-truth summary. In DUC2003 and DUC2004, the generated summary is truncated to 75 characters.

\subsection{Implementation Details}
\label{sec:impl}
We use the implementation from Huggingface for BART and T5. The models start from BART$_\text{LARGE}$ and T5$_\text{LARGE}$ checkpoints and are then further pre-trained for 1 epoch on our data\footnote{We found that more pre-training with lead bias does not bring additional gain.} which takes 47 hours on 32 V-100 GPUs. The batch size is 1,024. We use RAdam \cite{radam} as the optimizer, with a learning rate of $3\times10^{-4}$. Other hyper-parameters follow the settings in Huggingface. For example, for both BART-LB and T5-LB, the dropout rate is 0.1 and each input token is represented by a 1024-dim vector. BART-LB has 12 transformer layers and 16 attention heads in both the encoder and decoder, with 406M parameters. T5-LB has 24 transformer layers and 16 attention heads in both the encoder and decoder, with 770M parameters. 

In downstream summarization datasets, we employ the commonly used hyper-parameters from previous models (e.g. set in the configuration files of BART and T5) on the corresponding tasks. These hyper-parameters are for decoding based on beam search. Table~\ref{tab:hyperparam} shows the minimum summary length, maximum summary length and beam width for each task. The first two hyper-parameters decide the minimum / maximum number of tokens the decoder should generate. The beam width specifies the size of beam during beam search.

\begin{table*}[!h]
\centering
    \begin{tabular}{l|ccc|ccc|ccc} \thickhline
        \rowcolor[gray]{0.95} \textbf{Model} & \multicolumn{3}{c|}{\textbf{DUC2003}} & \multicolumn{3}{c|}{\textbf{DUC2004}} &
        \multicolumn{3}{c}{\textbf{XSum}}\\ 
        \thickhline
         & \textbf{R-1} & \textbf{R-2} & \textbf{R-L} & \textbf{R-1} & \textbf{R-2} & \textbf{R-L} & \textbf{R-1} & \textbf{R-2} & \textbf{R-L}\\
      \hline
      Lead & 21.30 & 6.38 & 18.82 & 20.91 & 5.52 & 18.20 & 16.30 & 1.60 & 11.95 \\ 
      \hline
      \rowcolor[gray]{0.95}\multicolumn{10}{c}{Supervised}\\ 
      \hline
      ABS \cite{gigaword} & 28.48 & 8.91 & 23.97 &  28.18 & 8.49 & 23.81 & / & / & /\\
      PEGASUS$^\text{FT}$ \cite{pegasus} & / & / & / & / & / & / & 47.21 & 24.56 & 39.25 \\
      \hline
      \rowcolor[gray]{0.95}\multicolumn{10}{c}{Unsupervised}\\ 
      \hline
      SEQ$^3$ \cite{seq3} & 20.90 & 6.08 & 18.55 & 22.13 & 6.18 & 19.30 & / & / & /\\
      \hline
      \rowcolor[gray]{0.95}\multicolumn{10}{c}{Zero-shot}\\ 
      \hline
      PEGASUS \cite{pegasus} & / & / & / & / & / & / & 19.27 & 3.00 & 12.72\\
      BART$_\text{LARGE}$ \cite{bart}& 6.69 & 1.56 & 5.94 & 13.58 & 2.91 & 12.10 & 19.26 & 3.30 & 14.67\\
      T5$_\text{LARGE}$ \cite{t5} & 10.11 & 2.43 & 9.25 & 13.61 & 2.91 & 12.23 & 19.66 & 2.91 &  15.31\\
      BART-LB & \textbf{20.43} & \textbf{5.80} & \textbf{17.89} & \textbf{21.88} & \textbf{6.24} & \textbf{19.22} & \textbf{26.18} & \textbf{7.60} & \textbf{20.92} \\
      T5-LB & 20.05 & 5.62 & 17.83 & 21.22 & 5.92 & 18.74 & 26.06 & 6.77 & 20.47\\

      \thickhline    
        \rowcolor[gray]{0.95} \textbf{Model} & \multicolumn{3}{c|}{\textbf{CNN/DM}} & \multicolumn{3}{c|}{\textbf{NYT}} & \multicolumn{3}{c}{\textbf{Gigaword}}\\
        \thickhline
      & \textbf{R-1} & \textbf{R-2} & \textbf{R-L} & \textbf{R-1} & \textbf{R-2} & \textbf{R-L} & \textbf{R-1} & \textbf{R-2} & \textbf{R-L}\\
      \hline
      Lead & 40.34 & 17.70 & 36.57 & 39.58 & 20.11 & 35.78 & 21.86 & 7.66 & 20.45 \\ 
      \hline
      \rowcolor[gray]{0.95}\multicolumn{10}{c}{Supervised}\\ 
      \hline
      BERTSUM \cite{bertsum} & 43.85 & 20.34 & 39.90 & 49.02 & 31.02 & 45.55 & / & / & / \\
      PEGASUS$^\text{FT}$ \cite{pegasus}& 44.17 & 21.47 & 41.11 & / & / & / & 39.12 & 19.86 & 36.24 \\
      \hline
      \rowcolor[gray]{0.95}\multicolumn{10}{c}{Unsupervised}\\ 
      \hline
      SEQ$^3$ \cite{seq3} & 23.24 & 7.10 & 22.15 & 17.85 & 3.94 & 19.53 & 25.39 & 8.21 & 22.68\\
      Brief \cite{wang2018learning} & 28.11 & 9.97 & 25.41 & / & / & / & 21.26 & 5.60 & 18.89\\
      TED \cite{ted} & 38.73 & 16.84 & 35.40 & / & / & / & 25.58 & 8.94 & 22.83\\
      \hline
      \rowcolor[gray]{0.95}\multicolumn{10}{c}{Zero-shot}\\ 
      \hline
      GPT-2 \cite{gpt} & 29.34 & 8.27 & 26.58 & / & / & / & / & / & /   \\
      PEGASUS \cite{pegasus}& 32.90 & 13.28 & 29.38 & / & / & / & 23.39 & 7.59 & 20.20\\
      BART$_\text{LARGE}$ \cite{bart} & 32.83 & 13.30 & 29.64 & 32.18 & 13.90 & 28.67 & 22.07 & 7.47 & 20.02\\
      T5$_\text{LARGE}$ \cite{t5} & 39.68 & 17.24 & 36.28 & 32.78 & 14.91 & 29.91  & 15.67 & 4.86 & 14.38\\
      BART-LB & \textbf{40.52} & \textbf{17.63} & \textbf{36.76} & 37.41 & 19.60 & 33.99 & \textbf{25.14} & \textbf{8.72} & \textbf{22.35} \\
      T5-LB & 38.47 & 16.62 &  35.23 & \textbf{40.27} & \textbf{20.81} & \textbf{36.88} & 24.00 & 8.19 & 21.62\\
      \thickhline       
    \end{tabular}
\caption{\label{tab:res} Results on the test set of all datasets, including ROUGE-1, ROUGE-2 and ROUGE-L. We use ROUGE recall scores in NYT and F1 scores in all other datasets. The highest scores in zero-shot are marked in bold.}
\end{table*}

\subsection{Baselines}
We include state-of-the-art models from several categories of summarization models. The supervised models include ABS \cite{gigaword} based on neural attention, BERTSUM \cite{bertsum} based on BERT \cite{bert} and the fine-tuned version of PEGASUS \cite{pegasus}, denoted by PEGASUS$^\text{FT}$. 
The unsupervised models include SEQ$^3$ \cite{seq3} based on article reconstruction, Brief \cite{wang2018learning} based on adversarial training and reinforcement learning, and TED \cite{ted} based on lead bias, theme modeling and denoising auto-encoder. The zero-shot baselines include large-scale pre-trained NLG models including PEGASUS \cite{pegasus}, GPT-2 \cite{gpt}, BART$_\text{LARGE}$ \cite{bart} and T5$_\text{LARGE}$ \cite{t5}.
We evaluate all zero-shot models on their pre-trained checkpoints without fine-tuning. We also include the extractive baseline of leading sentences. We follow the previous approaches \cite{seq3,tconvs2s} to use Lead-8 for Gigaword, Lead-1 for XSum, leading 75 characters for DUC2003/DUC2004 and Lead-3 for all the other datasets.

\subsection{Results}
Table~\ref{tab:res} show the results on all datasets, based on which we make the following observations.

Firstly, BART-LB and T5-LB outperform all zero-shot baselines in each dataset and achieve the best results among all non-supervised models in XSum, CNN/DailyMail and NYT. 
For instance, BART-LB outperforms the zero-shot version of PEGASUS by 7.6\%, 6.9\% and 1.8\% in ROUGE-1 on CNN/DailyMail, XSum and Gigawords, respectively. 

Secondly, the lead-bias pre-training dramatically improves the performance of the underlying generalized pre-trained model. For example, BART-LB outperforms BART$_\text{LARGE}$ by 13.7\%, 8.3\% and 6.9\% in ROUGE-1 on DUC2003, DUC2004 and XSum, respectively. T5-LB outperforms T5$_\text{LARGE}$ by 9.9\%, 7.6\%, 6.4\% in ROUGE-1 on DUC2003, DUC2004 and XSum, respectively.


Thirdly, BART-LB and T5-LB significantly outperform all unsupervised baselines in CNN/DM and NYT, and achieve very close results on the other datasets. We argue that although unsupervised models are fine-tuned with articles from target tasks, zero-shot models pre-trained on massive data can achieve similar or better results. Moreover, while an unsupervised model needs to be fine-tuned on downstream datasets, a single zero-shot model can be directly applied to many news summarization datasets, which makes deployment much more convenient.

Last, we note that although the pre-training is based on lead bias, the resulting model does not learn to simply copy leading sentences. In fact, BART-LB and T5-LB outperform the Lead baseline in 5 out of 6 datasets, ranging from 0.18\% to 9.9\% higher ROUGE-1 points. We conduct more investigations into this aspect in the following section.

\subsection{Insights}
\begin{figure}[t]
\centering
  \includegraphics[width=8.0cm]{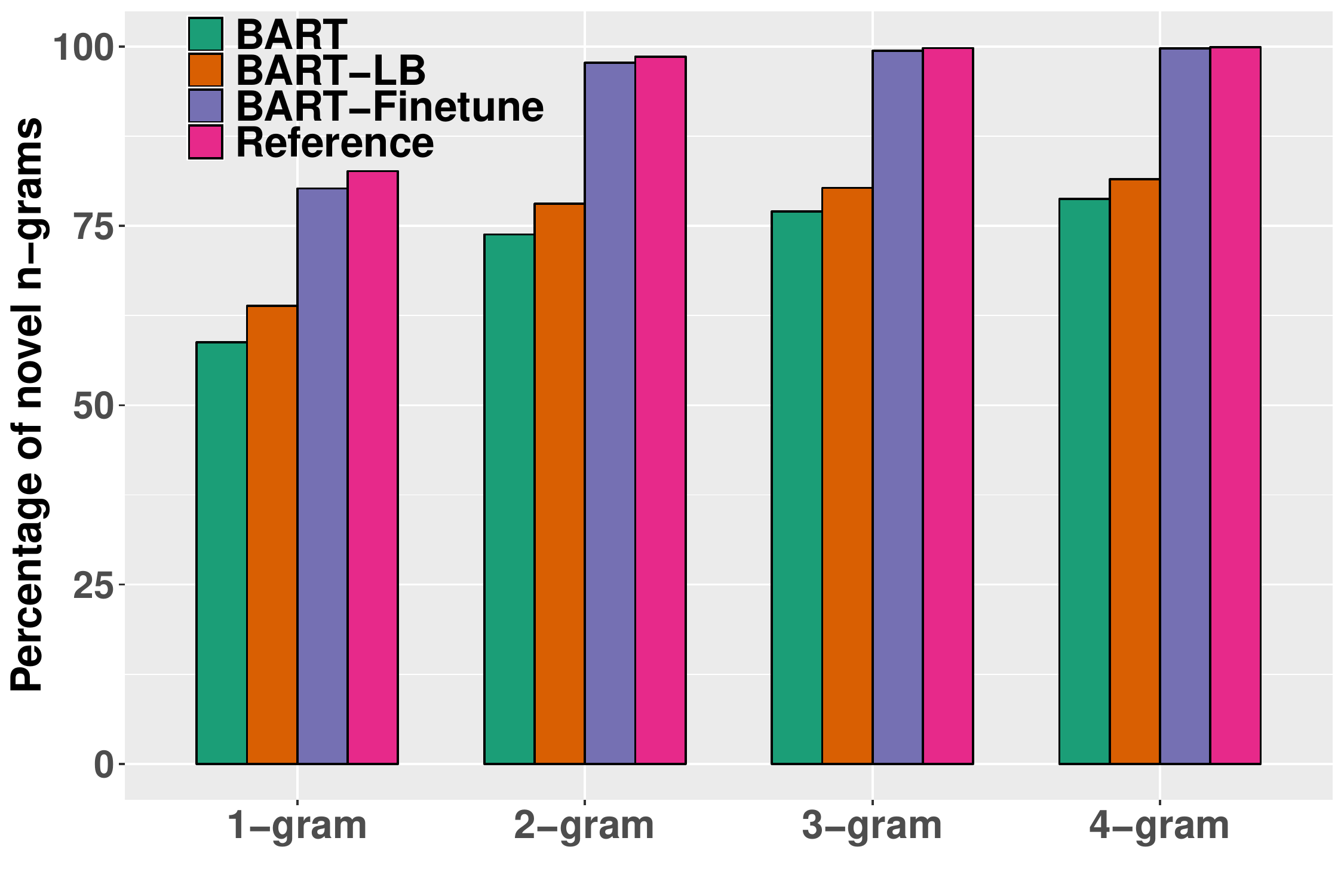}
  \caption{Ratio of novel n-grams, i.e. not in the article's leading sentence, in summaries from BART, BART-LB, BART-Finetune and the reference summary in XSum's test set.}
  \label{fig:novel}
\end{figure}

\textbf{Does our model simply copy leading sentences?} Since extracting leading sentences from articles as the summary can achieve high ROUGE scores in a number of news summarization datasets, we test whether our pre-trained model simply learns to copy the leading sentences. 

Following \cite{pgnet}, we compute the ratio of novel $n$-grams appearing in a model's summary that are not in the leading sentences. A higher ratio indicates that the model is more inclined not to copy the leading sentences from the article.

As shown in Figure~\ref{fig:novel}, the lead-bias pre-training increases the ratio of novel $n$-grams in XSum when we compare the results of BART and BART-LB. The reason is that during pre-training the model needs to predict LEAD-3 using \textit{the rest} of the article. Simply copying the first few sentences from Rest typically do not match LEAD-3. Thus, our proposed pre-training enforces that the model learns to comprehend and extract salient information from the whole article. And fine-tuning the pre-trained model on labeled data (BART-Finetune) can further increase the ratio of novel $n$-grams to a level close to reference summaries. 

\textbf{Effects of summary length.} We investigate whether the improvement brought by our lead-bias pre-training is affected by the length of summaries. Thus, we take the BART$_\text{LARGE}$ and BART-LB models, and compute the gain of BART-LB in ROUGE-1 score when the length of reference summary falls into different percentiles of the dataset: 0-20\%, 20-40\%, 40-60\%, 60-80\% and 80-100\%.

\begin{figure*}[h]
\centering
  \includegraphics[width=17cm]{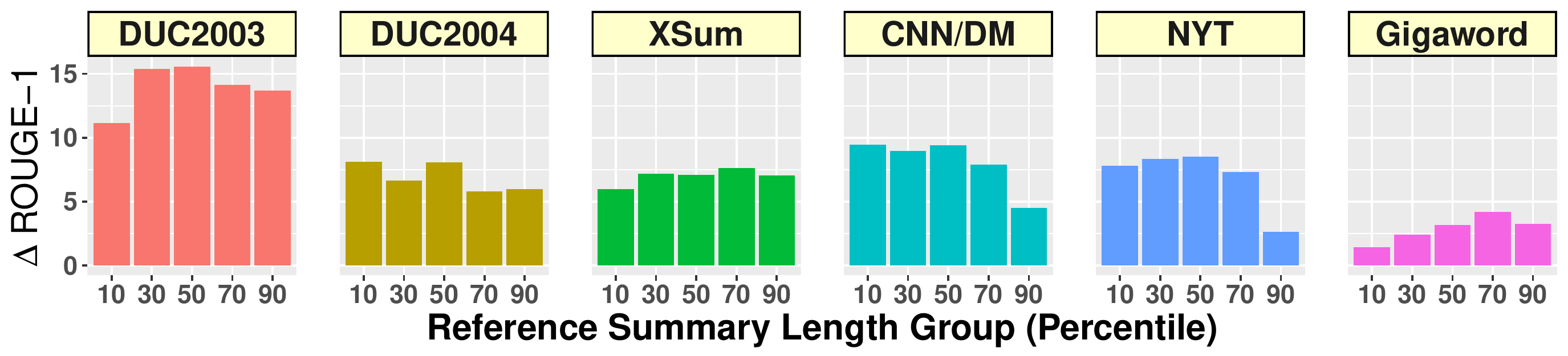}
  \caption{Averaged difference in ROUGE-1 score between the summaries from BART and BART-LB, grouped by the length of reference summary. For example, the first bin refers to the 0-20\% percentile of articles with the shortest reference summary in the corresponding dataset.}
  \label{fig:deltarouge}
\end{figure*}

\begin{figure*}[h]
  \begin{subfigure}{0.5\textwidth}
     \centering
     \includegraphics[width=.9\linewidth]{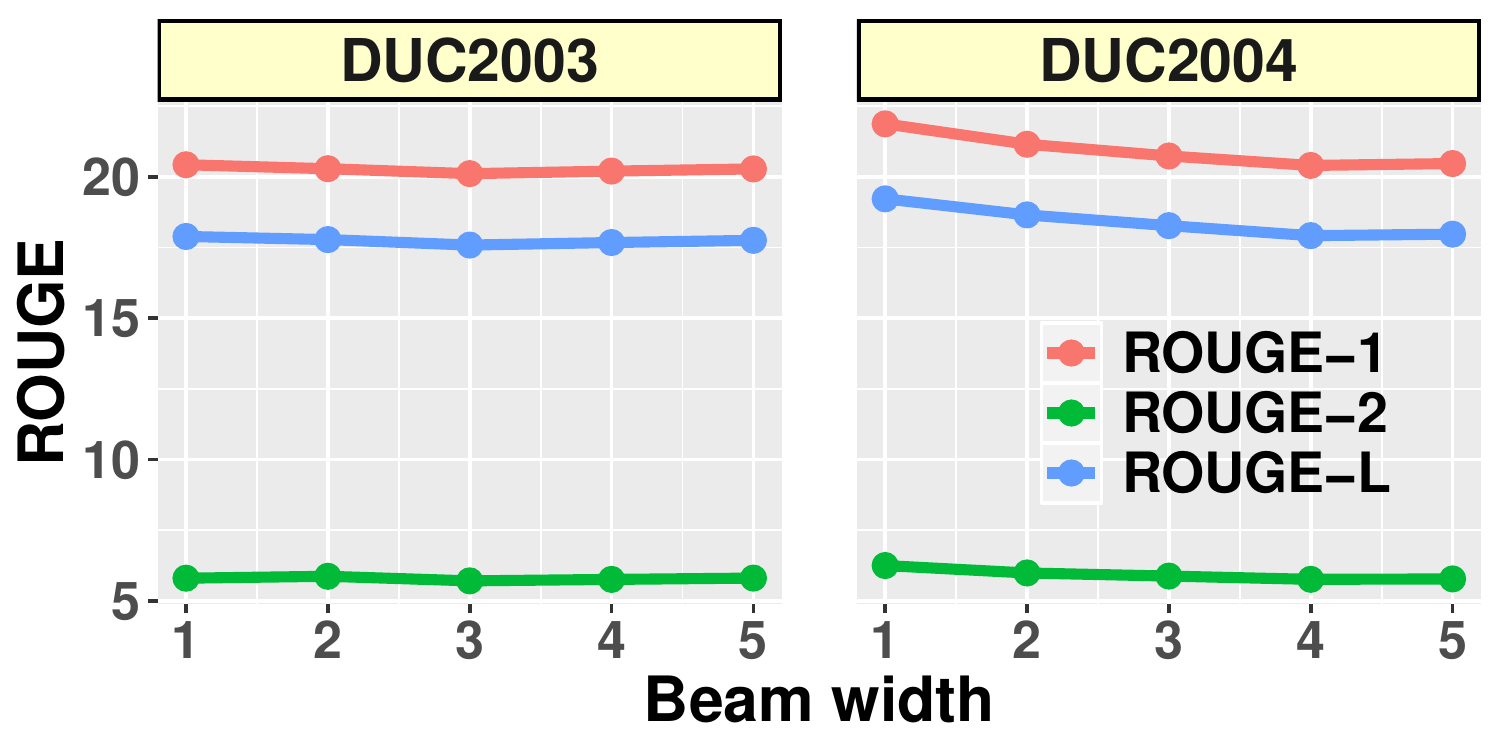}
  \end{subfigure}
  ~
  \begin{subfigure}{0.5\textwidth}
     \centering
     \includegraphics[width=.9\linewidth]{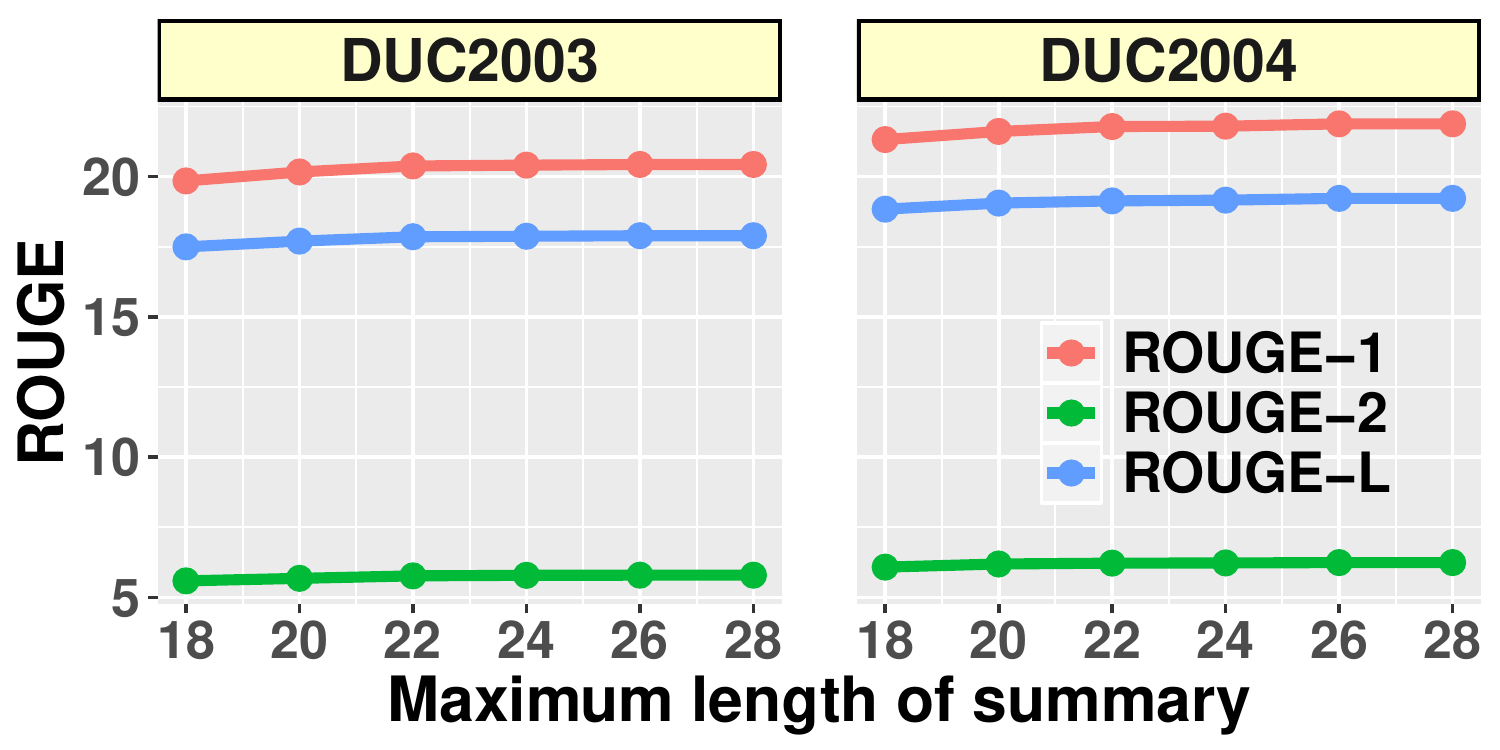}
  \end{subfigure}
  \caption{ROUGE scores for BART-LB on DUC2003 and DUC2004 under different hyper-parameters: beam width and maximum length of summary. Other hyper-parameters are set according to Table~\ref{tab:hyperparam}.}
  \label{fig:sensitivity}
\end{figure*}

As shown by Figure~\ref{fig:deltarouge}, in general, the gain of BART-LB over BART$_\text{LARGE}$ is the largest 
when the reference summary is at 40-60\% and 60-80\% percentile. Although longer summaries are typically more difficult to generate, the results indicate that our pre-training scheme can help more with producing better summaries with medium and medium-long lengths.

Meanwhile, we attribute the dip at 80-100\% to the capacity limit of the summarization model, as it becomes harder to achieve a high ROUGE score with a very long reference summary. This problem can be alleviated by developing more powerful generation models to produce long and consistent text.

\textbf{Sensitivity to hyper-parameters.} As our models are zero-shot, the hyper-parameters only apply to the decoding process, e.g. beam search width, minimum/maximum length of the decoded summary. We evaluate whether the quality of produced summaries is significantly affected by these hyper-parameters. 

Figure~\ref{fig:sensitivity} shows the performance of BART-LB on DUC2003 and DUC2004 under different beam width and maximum length of summary. Other hyper-parameters are set according to Table~\ref{tab:hyperparam}. As shown, most curves are flat, reflecting that our model is not very sensitive to the choice of hyper-parameters. 

\subsection{Human Evaluation}
We perform human evaluation to make sure that the gains achieved by our models in automatic evaluation align with human readability and quality.

We randomly sample 100 articles from the test set of DUC2003 and XSum. We then show the article, the reference summary and summaries generated by different models to the labelers. We randomly shuffle the order of summaries to reduce bias. The labelers need to judge the readability, i.e. how well the summary is written, and the relevance, i.e. how well the summary expresses the salient information from the article. For each category, the labeler needs to give an integer score between 1 and 5, with 5 being perfect. 

Each article-summary pair is judged by 3 labelers from Amazon Mechanical Turk and we show the average scores. To ensure labeling quality, we choose labelers with high approval ratings (>98\%).

As shown in Table~\ref{table:human}, although the Lead baseline has high readability scores due to direct copying, BART-LB and T5-LB outperform it in relevance. For example, in XSum, labelers rate our models 1 point higher on average than the Lead baseline in relevance.

Furthermore, the lead-bias pre-training can significantly improve both readability and relevance of the produced summaries (BART$_{\text{LARGE}}$ vs BART-LB and T5$_{\text{LARGE}}$ vs T5-LB). We argue that although BART and T5 have been trained on large-scale datasets, it is not specifically optimized for the summarization task, which requires a succinct recapitulation of the main points of the input article. As a result, a further pre-training in the news domain leveraging the lead bias can effectively improve the quality of summaries.

Finally, the reference summaries, which represent human-level summarization, achieve the highest readability and relevance scores. While there still exist gaps between summaries generated by models and humans, effective pre-training can help reduce the gap.

\begin{table}[tbp]
\centering
\begin{tabular}{l|c|c|c|c}
\thickhline
\textbf{Dataset} & \multicolumn{2}{c|}{\textbf{DUC2003}} & \multicolumn{2}{c}{\textbf{XSum}}\\
\hline
\textbf{Category} & \textbf{Read.} & \textbf{Rel.} & \textbf{Read.} & \textbf{Rel.}\\
\hline
Reference & 4.90 & 4.85 & 4.95 & 4.79 \\
\hline
Lead & \textbf{4.92} & 3.83 & \textbf{4.88} & 3.17\\
BART$_\text{LARGE}$ & 3.72 & 2.91 & 4.09 & 3.33\\
T5$_\text{LARGE}$ & 3.09 & 2.85 & 3.65 & 3.75\\
BART-LB & 4.27 & \textbf{3.95} & 4.10 & \textbf{4.25}\\
T5-LB & 4.39 & 3.78 & 4.02 & 4.12\\
\thickhline
\end{tabular}
\caption{Average scores (1-5) of readability (Read.) and relevance (Rel.) of summaries on DUC2003 and XSum's test sets. Each summary is judged by 3 human labelers.} 
\label{table:human}
\end{table}


\subsection{Qualitative analysis}
\begin{table*}[t]
    \centerline{
    \begin{tabular}{l|p{15cm}}
    \thickhline
        \textbf{Article} & A naturally red-headed sixth form student has been told to tone down her hair colour or she will be barred from classes for being `too ginger'. Emily Reay, 17, who has natural auburn hair but colours it so it is brighter, is a pupil at Trinity School in Carlisle, Cumbria, and claims she has sported the same vibrant ginger hair for the last three years. But on her return to lessons following the Easter holidays she claims she was ordered to tone it down after teachers decided the colour was inappropriate.
        Emily Reay (pictured) a pupil at Trinity School in Carlisle, claims she has sported the same colour of hair for the last three years. But now teachers have told her to tone it down or she will be banned from lessons. 
        `Everybody knows me as that ``young ginger singer"': Talented musician Emily says her hair is her trademark.
        Now, during her last year of A-levels, Emily - who is naturally auburn - has been told she is banned from lessons until she changes her appearance...
        \\
        \hline
        \textbf{Reference} & Sixth form student Emily Reay is barred from classes for being too ginger. Her school said colour was inappropriate and asked her to ``tone it down'' but she says her hair has been vibrant shade of red for over three years. Emily is a natural red head but colours her hair so it is even brighter.
        \\
        \hline
        \textbf{BART$_\text{LARGE}$} & AA pupil at Trinity School in Carn, where the school, and the colour,' the colour which, and whether, and how the colour and whether and how it, and if the colour of the colour in the tone and how, how the tone of the colours of the hair in the colours, how it is how the hair of a girl's hair, the way it affects how the age of a pupil's appearance, how much of a year of a-levels, whether it is the way a student's hair is a sign of how much a year is like a year. A naturally red-headed sixth form student has been told to tone down her hair colour.\\
        \hline
        \textbf{BART-LB} & A naturally red-headed sixth form student has been told to tone down her hair colour or she will be barred from classes for being `too ginger'. Emily Reay, 17, claims she has sported the same vibrant ginger hair for the last three years. But on her return to lessons following the Easter holidays she claims she was ordered to tone it down after teachers decided the colour was inappropriate.\\
        \hline
        \textbf{T5$_\text{LARGE}$} & Emily Reay, 17, a pupil at Trinity School in Carlisle, claims she has sported the same vibrant ginger hair for the last three years. But on her return to lessons following the Easter holidays she claims teachers have told her to tone it down or she will be barred from classes. Determined not to miss out on important lessons she will return to school with her hair pinned up. `I was very angry at first, and then burst into tears,' said\\
        \hline
        \textbf{T5-LB} & A naturally red-headed sixth form student has been told to tone down her hair colour or she will be barred from classes for being too ginger. Emily Reay, 17, who has natural auburn hair but colours it so it is brighter, is a pupil at Trinity school in Carlisle, Cumbria. She claims she has sported the same vibrant ginger hair for the last three years. Now, during her last year of A-levels, she has been banned from lessons until she changes her appearance. `I was very angry at first, and then burst into tears. I've\\
    \thickhline
    \end{tabular}}
    
    \caption{\label{tab:example} Reference summary and model-generated summaries for an example article in CNN/DailyMail dataset.
    }  
\end{table*}

\textbf{Example summaries.} We show an example article with summaries from the models and reference in Table~\ref{tab:example}. The article talks about a student Emily Reay who is barred from classes for her coloured hair. The reference summary includes the salient information in a logical order: the barring due to hair color, the reason and instruction given by the school and Emily's defense for her hair color. However, the summary from T5$_\text{LARGE}$ starts with Emily's defense, followed by the barring decision, making the logic flow non-straightforward. And it describes Emily's reactions, which is less important information in the article. BART$_\text{LARGE}$ also adopts the wrong order, as well as making many grammatical errors and containing duplicated phrases.

In comparison, both BART-LB and T5-LB produce summaries that contain the main points included in the reference summary in a logical order: the barring, the instruction given by the school and the student's defense for her hair color. Furthermore, both summaries rephrase and shorten the original article to make the statement more natural. For instance, the article sentence ``Emily Reay, 17, who has ..., and claims she has sported ...'' is restated as ``Emily Reay, 17, claims she has sported...'' in the summary of BART-LB, and as ``Emily Reay, 17, ..., Cumbria. She claims she has sported...'' in the summary of T5-LB. These are unique capabilities of abstractive summarization models compared with extractive methods.

\textbf{Error analysis.} We analyzed 50 articles from CNN/DailyMail and summaries generated by BART-LB. We found 23 summaries that contain various errors. We categorize the common types of errors in Table~\ref{tab:erroranalysis}. Note that one summary may contain multiple types of errors.

As shown, the most common type of error is the inclusion of insignificant information, which are often descriptions of event details. Invalid co-reference happens when a pronoun is copied from the article but the referent does not appear in the summary or appears in a different place. Incomplete ending of summary is usually caused by the maximum summary length specified in decoding. Spelling errors occur since both BART and T5 use sub-word tokenization which can sometimes lead to out-of-vocabulary words. Factual errors include distortion and fabrication, and it's a subject under increasing investigation for abstractive summarization \cite{zhu2020boosting}.

\begin{table}[h]
\centering
\begin{tabular}{l|l}
\thickhline
\textbf{Count} & \textbf{Type of error}\\
\hline
17 & Contains insignificant information from the article\\
\hline
12 & Invalid co-reference in the summary\\
\hline
10 & Incomplete sentences especially at the end\\
\hline
8 & Grammar \& spelling errors\\
\hline
5 & Factual errors, e.g. fabrication\\
\hline

\thickhline
\end{tabular}
\caption{Common types of errors found in the summary by BART-LB of 50 articles in CNN/DailyMail.} 
\label{tab:erroranalysis}
\vspace{-0.85cm}
\end{table}

\subsection{Deployment}
We have deployed the summarization model to Microsoft News to provide summaries to news articles. As the news come from various sources, we deploy the zero-shot summarization model which have been proven effective on multiple datasets.

As the summarization model is trained on English news, for a non-English article, we send it to Microsoft Azure Translator for translation into English, summarize it and translate the summary back to the original language.

We provide a demo website where one can paste news content to get the summary produced by the model. The summary is shown both in text and speech by Azure Text to Speech. The summarizer supports 30 different languages. Furthermore, we provide an API for developers to obtain summaries for news articles in different languages via POST request. More details about the demo website and API are in Appendix~\ref{sec:appendix_deployment}.

\section{Conclusions}
\label{Sec:conclusion}
In this paper, we propose a simple and effective pre-training method for zero-shot abstractive news summarization: treating the leading sentences from a news article as the target summary and the rest as the article. We collect a large-scale news corpus and conduct cleaning and filtering based on statistical analysis to ensure the quality of leading sentences as the delegate summary. We pre-train the generation model initialized with BART/T5 using the lead-bias data. The resulting models BART-LB and T5-LB outperform all zero-shot baselines and achieve comparable results with unsupervised methods on 6 benchmark datasets in both automatic and human evaluation.  Further analysis shows that our model does not simply copy leading sentences from the article and is insensitive to hyper-parameters. Finally, we deploy the model in Microsoft News and provide public APIs as well as a demo website for multi-lingual news summarization.
\bibliography{main}
\bibliographystyle{ACM-Reference-Format}

\appendix
\section{Deployment}
\label{sec:appendix_deployment}
\begin{figure*}[!h]
    \centering
    \includegraphics[trim=0cm 10cm 1cm 2cm, clip, width=0.75\linewidth]{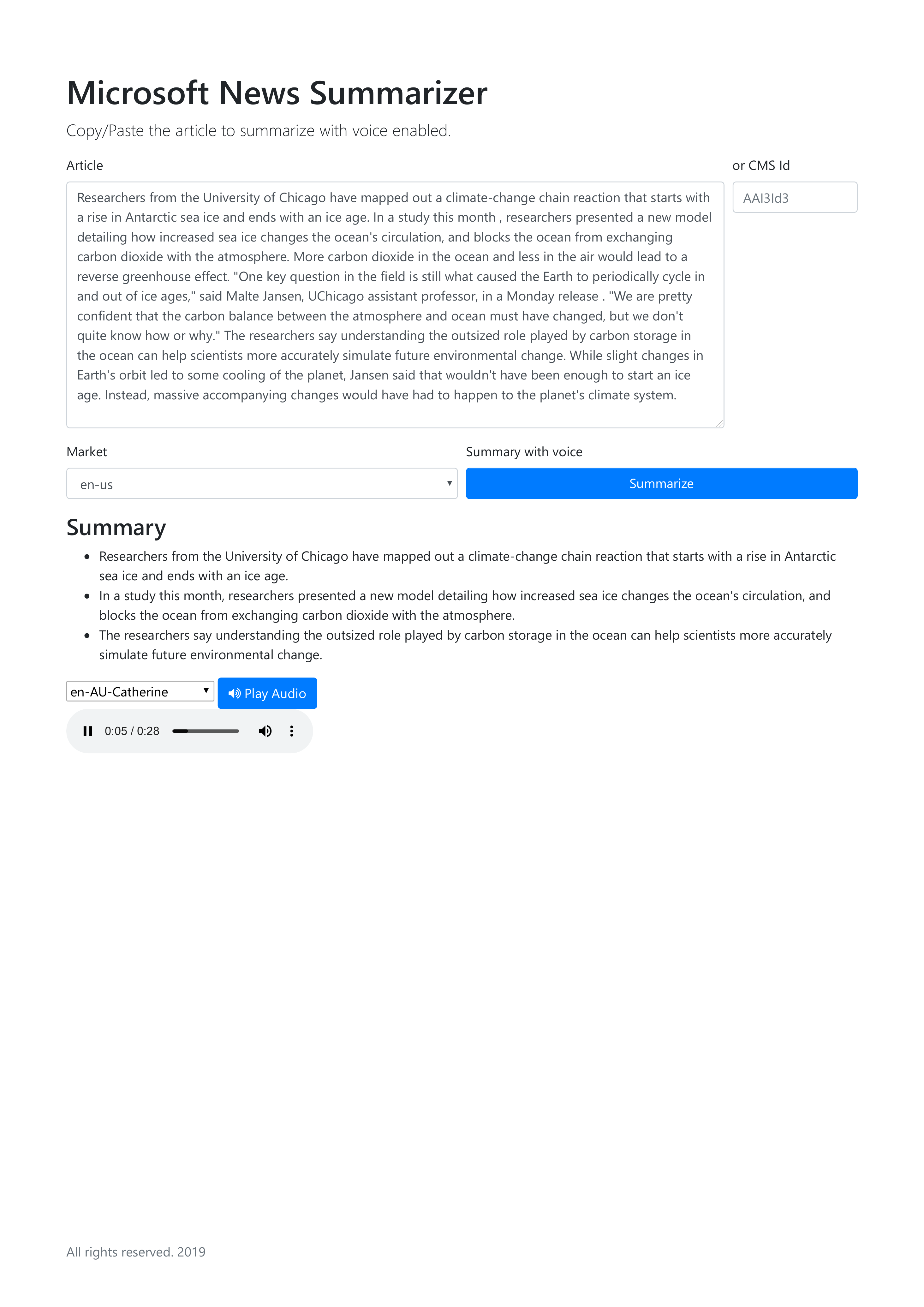}
    \vspace{-4cm}
    \caption{Demo website for our zero-shot summarization model. Via Microsoft Azure Translator, the model can summarize news articles from 30 different languages, which is returned both as text and speech by Azure Text to Speech. The demo website is at \url{https://msndata.azurewebsites.net/views/summary/index.html}.}
    \label{fig:demo}
\end{figure*}

We provide a demo website at \url{https://msndata.azurewebsites.net/views/summary/index.html} where one can paste news content to get the summary produced by our model. One can also use the CMS id\footnote{CMS is a unique id in the url for MSN news articles, e.g. the CMS id is \textit{BB19Hq5N} for the url \url{https://www.msn.com/en-us/news/politics/barack-obama-to-u-s-voters-our-elections-matter-to-everyone/ar-BB19Hq5N?ocid=Peregrine}.} to refer to MSN news articles. The summary is shown both in text and speech by Azure Text to Speech.
Although the model is trained on English news, we use Microsoft Azure Translator to translate non-English news to English, summarize it and then translate it back to the input language. The summarizer supports 30 different languages (Figure~\ref{fig:demo}). 

\begin{figure*}[!h]
    \centering
    \includegraphics[width=0.9\linewidth]{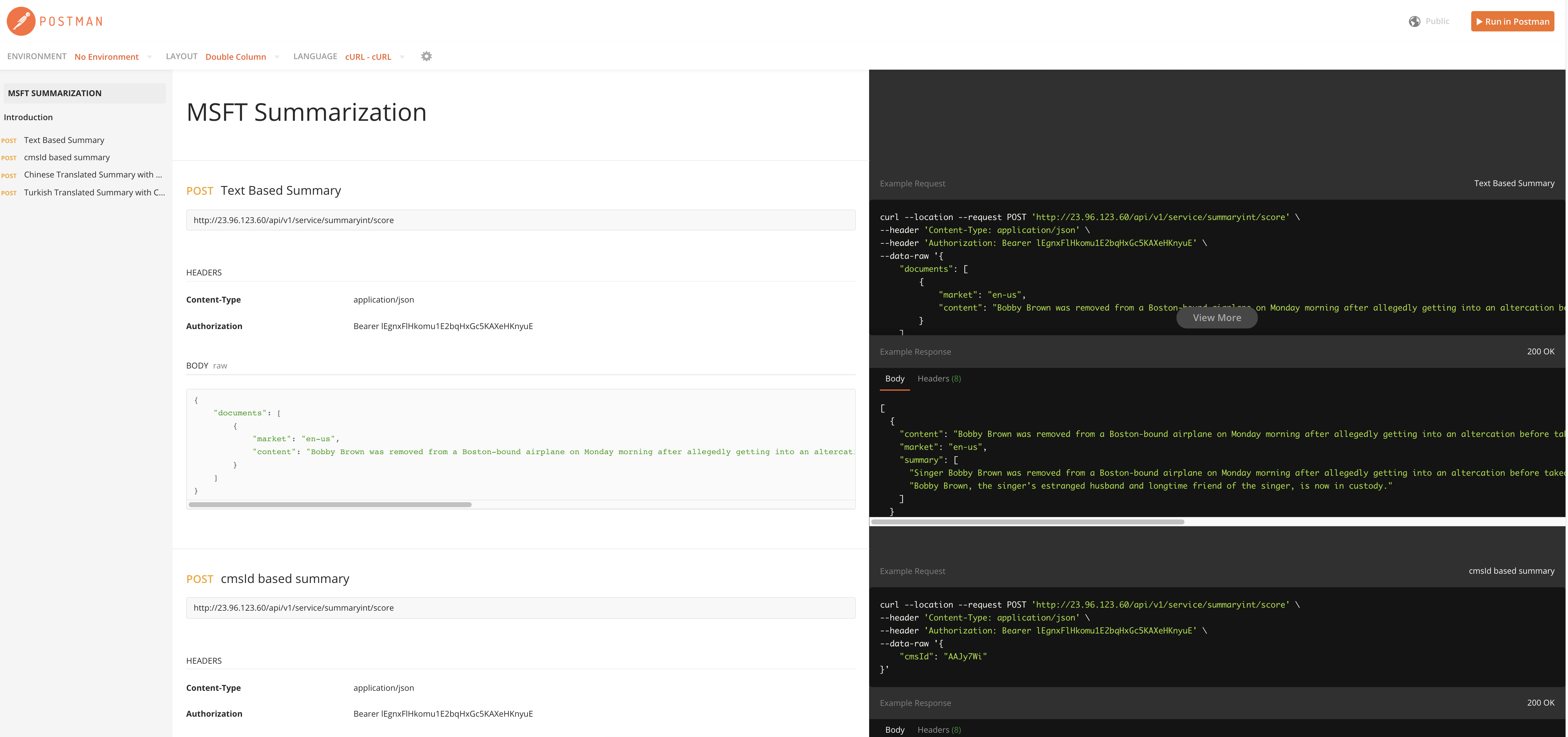}
    \caption{The interface of API for developers to obtain summaries for news articles in different languages using our model and Microsoft Azure Translator. The API is at \url{https://documenter.getpostman.com/view/4428744/SW7c2SXF?version=latest}.}
    \label{fig:api}
\end{figure*}

Furthermore, we provide an API for developers to obtain model-generated\footnote{We currently provide API to a model pre-trained on the same data from scratch, which has similar performance with BART-LB and T5-LB. We are preparing to release BART-LB and T5-LB soon.} summaries for news articles in different languages via POST request. Given an article and its language, the returned result will contain the model-generated summary in the corresponding language (Figure~\ref{fig:api}).

\end{document}